\documentclass{article}
\pdfoutput=1
\usepackage{spconf,amsmath,graphicx}
\usepackage[square,numbers]{natbib}
\usepackage{xcolor}
\usepackage[utf8]{inputenc} 

\usepackage[T1]{fontenc}    
\usepackage{hyperref}       
\usepackage{url}            
\usepackage{booktabs}       
\usepackage{amsfonts}       
\usepackage{nicefrac}       
\usepackage{microtype}      
\usepackage{threeparttable}
\usepackage{multirow}
\usepackage[normalem]{ulem}
\usepackage{xspace}
\usepackage{subcaption}
\usepackage{wrapfig}
\usepackage{ulem}
\usepackage{verbatim}
\usepackage{comment}

\title{Generating Annotated High-Fidelity Images containing Multiple Coherent Objects}
\name{Bryan Cardenas$^{\dagger}$, Devanshu Arya$^{\dagger}$ and Deepak K. Gupta$^{\dagger \star}$}
\address{$^{\dagger}$Informatics Institute, University of Amsterdam, The Netherlands\\
$^{\star}$Transmute AI Research, The Netherlands}
\begin{document}
%
\maketitle
\begin{abstract}
Recent developments related to generative models have enabled the generation of diverse and high-fidelity images. 
In particular, layout-to-image generation models have gained significant attention due to their capability to generate realistic and complex images containing distinct objects. 
These models are generally conditioned on either semantic layouts or textual descriptions. 
However, unlike natural images, providing auxiliary information can be extremely hard in domains such as biomedical imaging and remote sensing. 
In this work, we propose a multi-object generation framework\footnote{The code is available at \href{https://github.com/Cynetics/MSGNet/}{https://github.com/Cynetics/MSGNet/} } that can synthesize images with multiple objects without explicitly requiring their contextual information during the generation process. 
Based on a vector-quantized variational autoencoder (VQ-VAE) backbone, our model learns to preserve spatial coherency within an image as well as semantic coherency through the use of powerful autoregressive priors. 
An advantage of our approach is that the generated samples are accompanied by object-level annotations.
The efficacy of our approach is demonstrated through application on medical imaging datasets, where we show that augmenting the training set with the samples generated by our approach improves the performance of existing models.
\end{abstract}
\begin{keywords}
Generative Modelling, VQ-VAE, PixelSNAIL, Multi-Object Generation, Data Augmentation
\end{keywords}
\section{Introduction}
Recent advancements in deep learning-based generative modelling techniques have made it feasible to generate diverse high-fidelity images \cite{brock2019iclr, Oord2017neurips, Razavi2019iclr}.
Most of these methods focus on images with one centralized object (\emph{e.g.}, faces in CelebA \cite{Liu2015celebA} and objects in ImageNet \cite{Russakovsky2015ijcv}). However, most real-life natural images contain multiple interrelated objects distributed across the image (\emph{e.g}., MS-COCO dataset \cite{Tsung-yi2014eccv}). To generate such images, generative models should be able to jointly learn the complex object-object and object-background relationships. This requires explicit control over the kind of objects generated, their shapes as well as their locations in the image. 

Few approaches have recently addressed the above issue by conditioning the generative process with textual descriptions, these include using natural language descriptions ~\cite{reed2016icml,zhang2017iccv}, image category labels ~\cite{miyato2018iclr,zhang2019icml} or scene graphs ~\cite{johnson2018cvpr, yikang2019nips}. Other works have considered using semantic layout as auxiliary information to further control the overall placement of objects in the image~\cite{wang2018cvpr,Hinz2019iclr, spade}. However, engineering these additional inputs adds significant extra effort on the generation process, and the diversity of the generated images is confined by the quality of auxiliary information.
We also observe that this is yet predominantly done with GAN-based models introducing problems such as mode collapse, unstable training or a lack of diversity \cite{Oord2017neurips}.
Furthermore, it is to be noted that information such as textual description or semantic layout are attainable for simple \textit{natural images} that can easily be interpreted by annotators. By natural images, we refer to images comprising common objects such as cars or dogs. Other images such as medical images or satellite images will be referred to as \emph{non-natural images}. Most such images describe complex processes, and using simple captions to explain the inherent object-object interactions can be extremely hard for even the domain experts. We argue that manually providing semantically coherent layouts for such images would be more difficult; there is a need for an approach that can \emph{implicitly} capture the inherent coherency among various objects distributed across any image, as well as the spatial coherency among these objects.

In this paper, we introduce Multi-Object Semantic Generation Network (MSGNet) that addresses the challenges outlined above. MSGNet can generate images containing multiple coherent objects without explicitly requiring any contextual information during the generation process. Our approach uses an adapted Vector Quantized Variational Autoencoder (VQ-VAE, \cite{Oord2017neurips}), and is conditioned with two strong autoregressive priors based on PixelSNAIL \cite{chen2017pixelsnail}.  At generation time, we synthesize images together with annotations, thereby eliminating the added annotation burden. Samples generated by MSGNet are realistic and diverse, thus, it can be used as a data augmentation technique for problems related to natural as well as non-natural images. To study the working of MSGNet, we perform here a series of experiments on the CLEVR dataset \cite{johnson2017clevr} and we investigate its applicability as a data augmentation technique for segmentation tasks in medical imaging. 
\begin{figure*}[h!]
    \centering
    \includegraphics[width = 0.8\textwidth]{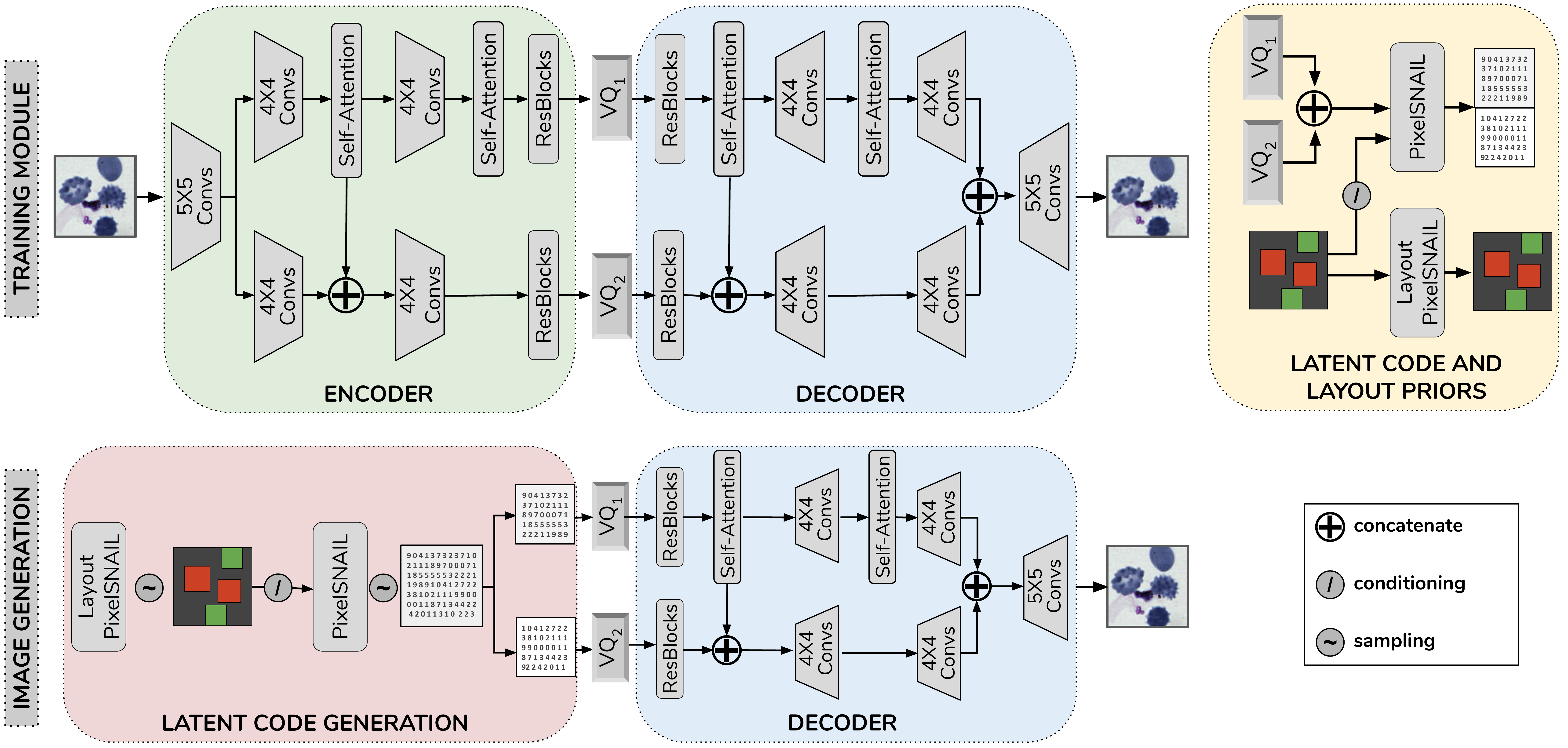}
    \caption{Schematic representation of Multi-Object Semantic Generation (MSGNet) framework. The double-path VQ-VAE model learns to reconstruct an image by first encoding it to a discrete latent. After training the VQ-VAE, a PixelSNAIL model learns the discrete latent distribution, while the LayoutPixelSNAIL learns the semantic distribution of the layout. }
    \label{fig:architecture}
\end{figure*}


\section{Approach}
We use a VQ-VAE model \cite{Oord2017neurips} that encodes an image $\mathbf{x}$ to an embedding $E(\mathbf{x})$, subsequently the embedding is quantized and then reconstructed by the decoder network. More formally, each element of $E(\mathbf{x})$ is assigned an index $i$ that corresponds to a codebook prototype-vector $\mathbf{e}_i \in \{\mathbf{e}_0,\dots,\mathbf{e}_K$\}, where $K$ is the size of the codebook. In the vector quantizing step, the index is assigned by considering its nearest prototype-vector as: $i = \text{argmin}_j \lVert E(\mathbf{x}) - \mathbf{e}_{j}\rVert$.
%
When decoding, these indices are used to obtain the corresponding prototype-vectors from the codebook and reconstruct input $\mathbf{x}$ using the decoder $D(\mathbf{e})$. Since the vector quantization step is non-differentiable, the gradient of the error is back-propagated to the encoder by using the straight-through gradient estimator \cite{ste2019}. Generally, VQ-VAE models are trained with a three-part loss function:  
    $\mathcal{L}_{total} = \mathcal{L}_{rec}(\mathbf{x}, D(\mathbf{e})) + \mathcal{L}_{cb}(\mathbf{e}, E(\mathbf{x})) + \mathcal{L}_{c}(\mathbf{e}, E(\mathbf{x}))$.
In this study, we use the mean squared error for $\mathcal{L}_{rec}$. The codebook loss $\mathcal{L}_{cb}$ forces the prototype-vectors to draw closer to the output of the encoder $E(\mathbf{x})$, while the commitment loss $\mathcal{L}_{c}$ ensures that $E(\mathbf{x})$ commits to a prototype-vector. For further related details, we refer the reader to \cite{Oord2017neurips}.

Our proposed Multi-Object Semantic Generation Network (MSGNet, Fig. \ref{fig:architecture}) maintains spatial and semantic coherency within an image through two strong autoregressive priors based on PixelSNAIL \cite{chen2017pixelsnail}. The first PixelSNAIL learns the distribution of the latent encodings of an image, while the other is used to specifically learn the semantic distribution (layout) of an image. The PixelSNAIL model factors the joint distribution as a product of parameterized conditionals: $p_{\mathbf{\theta}}(\mathbf{x}) = \prod_{i=0}^{n} p_{\mathbf{\theta}}(x_{i}\lvert \mathbf{x}_{<i}, \mathbf{h}_{<i})$, where $\mathbf{h}$ is used to condition the model further. An implicit advantage of our approach is that the generated samples are accompanied by their corresponding layout.  

\begin{figure}[h!]
\centering
\begin{subfigure}{0.34\textwidth}
\centering
\includegraphics[scale=0.85]{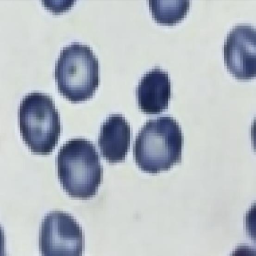}
\includegraphics[scale=0.85]{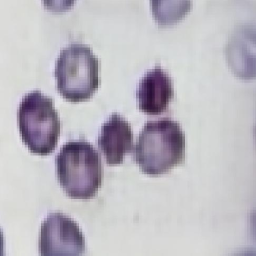}
\includegraphics[scale=0.85]{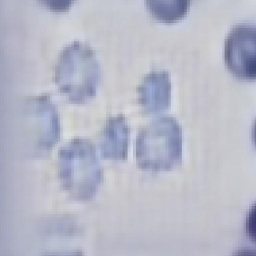}
\end{subfigure}
\caption{
Reconstructed cell image (left) and the contributions obtained independently from the self-attention path (middle) and the normal path (right) in MSGNet.
}
\label{fig_complementary}
\end{figure}

Another benefit of MSGNet is a more efficient use of multiple discrete latent codebooks, which enables us to use one PixelSNAIL model. In MSGNet, the VQ-VAE module uses two codebooks, $VQ_{1}$ and $VQ_{2}$, of equal dimensions to reconstruct an image using complementary features. Fig. \ref{fig_complementary} demonstrates how each path contributes towards the reconstruction of the original image. 



\textbf{Encoder and Decoder}. As shown in Fig. \ref{fig:architecture}, MSGNet uses a two-path approach where one path has self-attention modules \cite{zhang2019icml, vaswani2017att} responsible for learning features and dependencies from distant parts in an image while the second path solely uses regular convolutions. 

\begin{figure*}[h!]
\centering
\begin{subfigure}{\textwidth}
\centering
\includegraphics[scale=0.9]{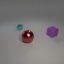}\hspace{0.8em}
\includegraphics[scale=0.9]{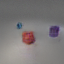}
\includegraphics[scale=0.9]{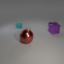}
\includegraphics[scale=0.9]{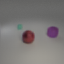}
\includegraphics[scale=0.9]{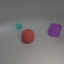}
\includegraphics[scale=0.9]{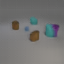}

\end{subfigure}

\begin{subfigure}{\textwidth}
\centering
\includegraphics[scale=0.9]{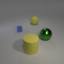}\hspace{0.8em}
\includegraphics[scale=0.9]{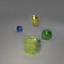}
\includegraphics[scale=0.9]{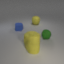}
\includegraphics[scale=0.9]{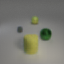}
\includegraphics[scale=0.9]{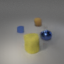}
\includegraphics[scale=0.9]{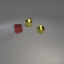}

\end{subfigure}

\caption{Two samples using CLEVR. The original image (left-most column), and generations obtained using obj-cGAN, pre-MSGNet, Pix2Pix, No-Comp MSGNet, and BigGan (right-most column). BigGAN does not take any label information.
}

\label{fig_clevr_basic}
\end{figure*}

\textbf{Latent code and layout priors}. After training the enhanced VQ-VAE component, the two latent codes of size $H\times W$ are extracted from the codebooks ${VQ_1}$ and $VQ_2$. To avoid training two priors over the two latent codes, as done in \cite{Razavi2019iclr}, we concatenate these codes, resulting in a single $2H\times W$ shaped latent code. A PixelSNAIL model is used to learn the prior distribution over the concatenated latent codes. Furthermore, the layout that corresponds to the encoded image is bilinearly downsampled to a size of $H\times W$. This downsampled layout is used as the condition comprising the spatial and object-type information in the image. Since the layouts are usually less complex than the image latent codes, we used smaller PixelSNAIL models to learn the layouts to train and generate layouts quickly; we call this the LayoutPixelSNAIL. 

\textbf{Layout and latent code generation}. To generate new samples we first sample a layout from the LayoutPixelSNAIL. This layout is bilinearly upsampled and provided as conditioning to the PixelSNAIL that generates the latent codes. This generated latent code image is split back into two different latent codes corresponding to outputs from $VQ_1$ and $VQ_2$. These two latent codes are then decoded to reconstruct the image, the procedure is shown in the bottom part of Figure \ref{fig:architecture}.

\section{Evaluation and Analysis}
\subsection{Generating high-fidelity multi-object images}
 To perform quantitative comparison, we assume here that layout information, i.e., bounding boxes and labels, are explicitly provided to our model. We refer to this modified version of our approach as pre-MSGNet. Additionally, we compare pre-MSGNet to a version that lacks a second path, referred as No-Comp MSGNet, to test the justification for complementary features; using two paths with two codebooks. Further. We compare our results to four other methods. We choose BigGAN and Pix2Pix as baselines, where the latter was specifically designed for semantic generation, and the former deems as an a-semantic baseline. Our results are also contrasted with the recent multi-object generation method using a GAN-based model with object pathways \cite{Hinz2019iclr}, referred to as \emph{obj-cGAN}. 

We follow a similar setup as \cite{Hinz2019iclr} to evaluate MSGNet and the other methods. We use  a U-Net \cite{ronneberger2015u} to assess the performance of the generative process based on the improvement observed in $F_{1}$-score and FID \cite{fid}. This is done in two settings, first, after augmenting the training set with the generations and, second, when training the U-Net with only the generated samples. The second case measures the diversity of the generations, as seen in \cite{Ravuri2019ClassificationAS}. 

\begin{figure}[h]
\centering
\begin{subfigure}{0.49\textwidth}
\centering
\includegraphics[scale=0.3]{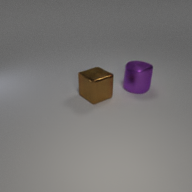}\hspace{-0.2em}
\includegraphics[scale=0.3]{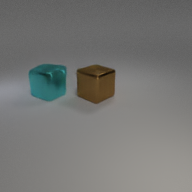}\hspace{-0.2em}
\includegraphics[scale=0.21]{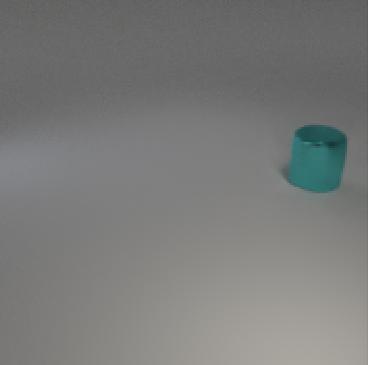}\hspace{-0.2em}
\includegraphics[scale=0.158]{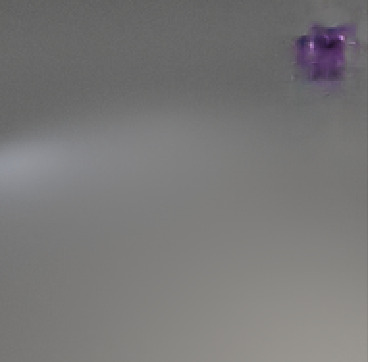}
\caption{Pre-MSGNet}
\end{subfigure}
\begin{subfigure}{0.49\textwidth}
\centering
\includegraphics[scale=0.3]{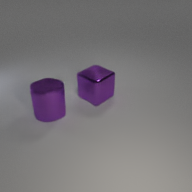}\hspace{-0.2em}
\includegraphics[scale=0.3]{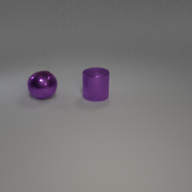}\hspace{-0.2em}
\includegraphics[scale=0.3]{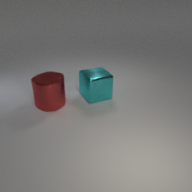}\hspace{-0.2em}
\includegraphics[scale=0.3]{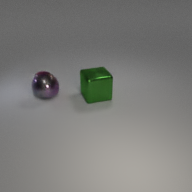}
\caption{MSGNet}
\end{subfigure}
\caption{The added constraint included one object positioned at the center and no object at the corners, and reflections should be coherent. MSGNet always generated correct examples.}
\label{fig_clevr_semantic}
\end{figure}

\begin{table}
  \caption{$F_{1}$-score scores measured for quality and diversity of the sample generations and FID. The $F_{1}$-score (accuracy, $A$) reflects a baseline for the pre-MSGNet performance on generated data with respect to the other baselines, while the $F_{1}$-score (diversity, $D$) shows how robust and diverse the generated data is. }
  \label{table-mnist-clevr}
  \centering
  \begin{threeparttable}[t]
  \centering
  \begin{tabular}{ l | l | l | l}
  \toprule
    Model & $F_{1}$-score (A) & $F_{1}$-score (D) & FID\\
    \midrule
    BigGAN \cite{biggan} & - & - & 122.3\\
    Pix2Pix\cite{pix2pix} & \textbf{0.745} & 0.731 & 99.1 \\
    No-Comp MSGNet & 0.711 & 0.797 & 65.7\\
    pre-MSGNet & 0.738 & \textbf{0.843} & \textbf{57.3}\\
    \emph{obj-cGAN} & 0.573 & 0.563 & 114.6\\

    \bottomrule
  \end{tabular}
  \end{threeparttable}
\end{table}
Table 1 shows how pre-MSGNet compares to other methods in terms of $F_{1}$ and FID scores. The $F_{1}$- score obtained using our approach is significantly higher than the other baselines on the CLEVR dataset. Similarly, MSGNet outperforms all the methods in terms of the diversity  of the generated samples. \mbox{Fig. \ref{fig_clevr_basic}} shows two examples of CLEVR generations with the considered methods. 
We further study the semantic and spatial coherency on the CLEVR dataset by imposing a constraint, where every image should contain one object centered in the image, and another object randomly positioned around it so that the reflections are visible. We observed that our MSGNet always generates reliable samples (see Fig. \ref{fig_clevr_semantic}).

\subsection{Application in Data Augmentation}\label{dataugment}
The primary goal behind designing MSGNet is to generate new \emph{image samples with object-level annotations}, especially for non-natural images.
We therefore test MSGNet on two medical problems: malaria cell detection \cite{Ljosa2012naturemethods} and brain tumor segmentation \cite{menze2014medicalimaging}. To assess the added value of data augmentation, we evaluate the performance of a U-Net \cite{ronneberger2015u} model trained with and without using additional generated samples in the training set and study the difference between the two $F_{1}$-scores.
\begin{figure}[h!]
    \centering
    \begin{subfigure}{0.49\textwidth}
    \centering
    \includegraphics[scale=0.48]{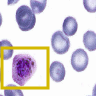}
    \includegraphics[scale=0.48]{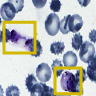}
    \includegraphics[scale=0.48]{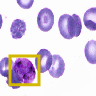}
    \includegraphics[scale=0.48]{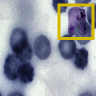}
    \caption{Real samples}
    \end{subfigure}
    \begin{subfigure}{0.49\textwidth}
    \centering
    \includegraphics[scale=0.48]{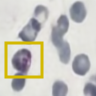}
    \includegraphics[scale=0.48]{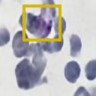}
    \includegraphics[scale=0.48]{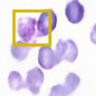}
    \includegraphics[scale=0.48]{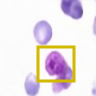}
    \caption{MSGNet generations}
    \end{subfigure}
    \caption{The bounding boxes mark the malaria infested cells.}
    \label{fig_mognet_malaria_samples}
\end{figure}
\textbf{Malaria cell detection.} The original dataset \cite{Ljosa2012naturemethods} comprises images of cells, either infected by malaria or not. The dataset was down-sampled to yield $96\times 96$ images containing cells.  
\begin{figure}
\begin{subfigure}{0.5\textwidth}
\centering
\includegraphics[scale=0.5]{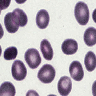}
\includegraphics[scale=0.5]{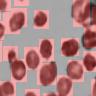}
\includegraphics[scale=0.5]{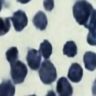}
\includegraphics[scale=0.5]{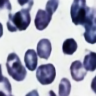}
\includegraphics[scale=0.5]{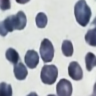}
\end{subfigure}
\begin{subfigure}{0.5\textwidth}
\centering
\includegraphics[scale=0.5]{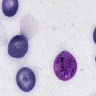}
 \includegraphics[scale=0.5]{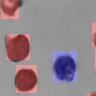}
\includegraphics[scale=0.5]{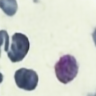}
\includegraphics[scale=0.5]{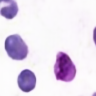}
\includegraphics[scale=0.5]{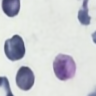}
\end{subfigure}
\caption{The first column shows the real patches, while the second one shows the overlay of the generations; Red and Blue show healthy and malaria infested cells respectively. The generations conditioned over the masks are shown in columns 2-5. The variation among the four samples demonstrates the potential of MSGNet to generate diverse samples.}
\label{malaria_examples_with_masks}
\end{figure}
Fig. \ref{fig_mognet_malaria_samples} shows four image samples of malaria-infested cells from the original dataset as well as four images generated from MSGNet. 
We also investigate the diversity qualitatively. We take masks from the training set, and condition the generations of MSGNet. The qualitative demonstration of diversity is shown in \mbox{Fig. \ref{malaria_examples_with_masks}}, we see that the generations differ from each other as well as from the original training sample. 
By augmenting the training set with MSGNet generated samples, we observe that MSGNet generations improve the $F1$-score by 2.2\% ($SD=0.008$) and by 1.1\% ($SD=0.006$) when an ImageNet pretrained DeseNet-161 encoder \cite{densnetgao} is used. We hypothesize that the prior information brought in by the powerful encoder allows the model to generalize better, and the scope of improvement left for augmentation is reduced.

\begin{figure}[h!]
\begin{subfigure}{0.48\textwidth}
\centering
\includegraphics[scale=0.26]{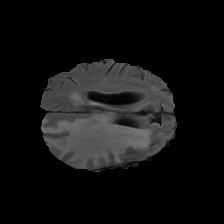}
\includegraphics[scale=0.26]{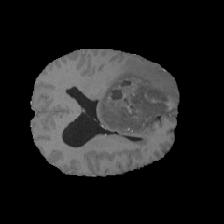}
\includegraphics[scale=0.26]{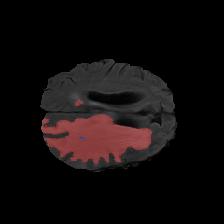}
\includegraphics[scale=0.26]{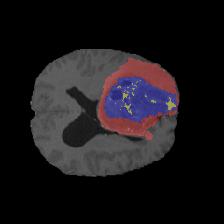}
\caption{Real samples}
\label{fig_brain_real}
\end{subfigure}
\hspace{1em}
\begin{subfigure}{0.48\textwidth}
\centering
\includegraphics[scale=0.26]{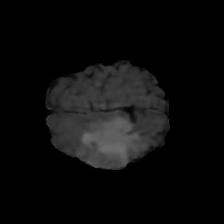}
\includegraphics[scale=0.26]{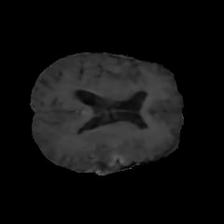}
\includegraphics[scale=0.26]{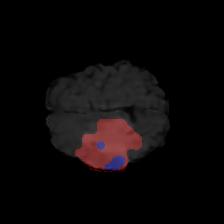}
\includegraphics[scale=0.26]{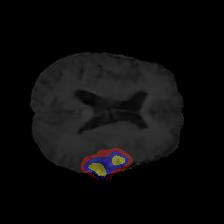}
\caption{MSGNet generations}
\end{subfigure}
\caption{Image slices of brain and the respective overlaid tumor masks, Red: peritumoral edema, Yellow: necrotic and non-enhancing tumor core, Blue: enhancing tumor.}
\label{fig_brain_gen}
\end{figure}
\textbf{Brain tumor segmentation.} Segmenting tumors in brain is challenging due to the high variance in appearance and shape of the tumors \cite{sun2019neuroscience}. 
We use a subset of the 2017 Brain Tumour Image Segmentation (BraTS) dataset ~\cite{bakas2017nature, menze2014medicalimaging}. It contains multi-parametric 3D MRI scans. The segmentation masks include (possibly overlapping) three classes of tumor. The 3D scans are split into 2D image slices, and only images that contain at least one tumor class are retained. 
Figure \ref{fig_brain_gen} shows three MSGNet generated samples as well as the corresponding masks. Since MSGNet adheres to the underlying relationships between various classes, we see that the generated images and the masks for different tumor classes are in line with the images and masks from the training set. In general, MSGNet rarely generated erroneous samples; we found roughly one sample per 3000 generations that could be rejected. Similar to the malaria experiment, we observed an improvement of 3.1\% ($SD=0.003$) in $F_{1}$-score on the validation set. 
\section{Conclusion}
We have introduced MSGNet: a generative negative log-likelihood-based framework that implicitly models the relationship between multiple objects within an image without requiring any auxiliary information at generation. This was useful with non-natural images such as in brain tumor segmentation. MSGNet generations suggest that they are of sufficient fidelity and diversity to augment biomedical imaging datasets.

\bibliographystyle{IEEEbib}
\bibliography{refs}
\section{Appendix}
\setcounter{page}{1}

\subsection{Datasets and Additional Results}
This section provides additional discussion related to the datasets (Multi-MNIST, CLEVR, Malaria and BraTS) used in this study as well as a few additional generations obtained using MSGNet.

\subsection{Multi-MNIST}\label{mmnist-appendix}
\begin{figure*}[h!]
    \centering
    \includegraphics[width = 0.75\textwidth]{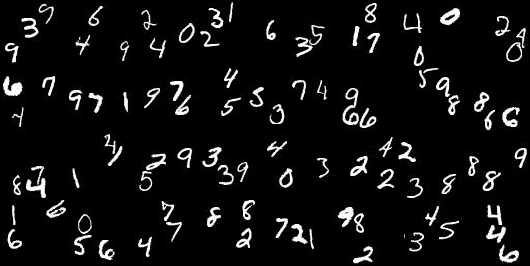}
    \caption{Original Multi-MNIST images. Each $64\times 64$ image consists of exactly three digits.}
    \label{fig:origmnist}
\end{figure*}

The Multi-MNIST dataset that has been used to study the generation fidelity of multi-object images is obtained from \cite{Hinz2019iclr}, which was adapted from \cite{eslami2016mnist}. The dataset consisted of $50,000$ images, each of dimension $64\times 64$, containing three MNIST digits with no overlap and with no background. The digits were placed at random locations.  \autoref{fig:origmnist} shows a set of real Multi-MNIST images whose bounding boxes have been used for the generation process. Note that controlling the location and object-type was done by providing the full layout of the image. That is, we used the bounding box information to fill the corresponding image area with the corresponding label. We constrained the dataset further to analyze how well MSGNet could learn the relationships between various objects in the spatial and semantic coherency experiments. Both experiments used images of the same size and the datasets contained the same number of images compared to the fidelity experiments. In the Multi-MNIST experiments overlapping digits were not allowed.

\begin{figure*}[h!]
    \centering
    \includegraphics[width = 0.75\textwidth]{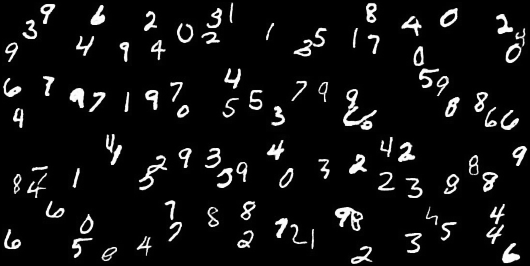}
    \caption{Generated Multi-MNIST images using the bounding boxes of \autoref{fig:origmnist} as reference.}
    \label{fig:genmnist}
\end{figure*}

\subsection{CLEVR}\label{clevr-appendix}
\begin{figure*}[h!]
    \centering
    \includegraphics[width = 0.75\textwidth]{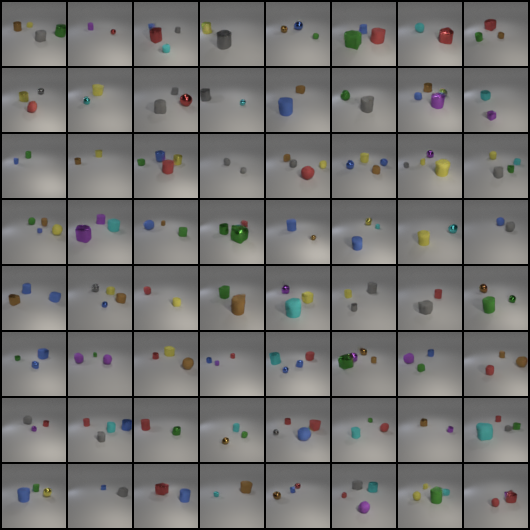}
    \caption{Generated $64\times 64$ CLEVR images.}
    \label{fig:64clevr}
\end{figure*}
The CLEVR dataset \cite{clevr2016} was originally designed for compositional language and visual reasoning, in this study we adopted the dataset as mentioned in \cite{Hinz2019iclr}, where only the geometric objects and their properties are taken into consideration. That is, we used 25,000 CLEVR images containing 2-4 objects, each of dimension $64 \times 64$. There were three different shapes: cylinders, spheres and cubes, and each can have eight different colours, making it a total of 24 possible classes. Thus, we conditioned the LayoutPixelSNAIL models on these 24 classes and one background class to have full control of the generation process. In \autoref{fig:64clevr}, we show 64 generated CLEVR images.

\begin{figure*}[h!] 
    \centering
    \includegraphics[scale=0.25]{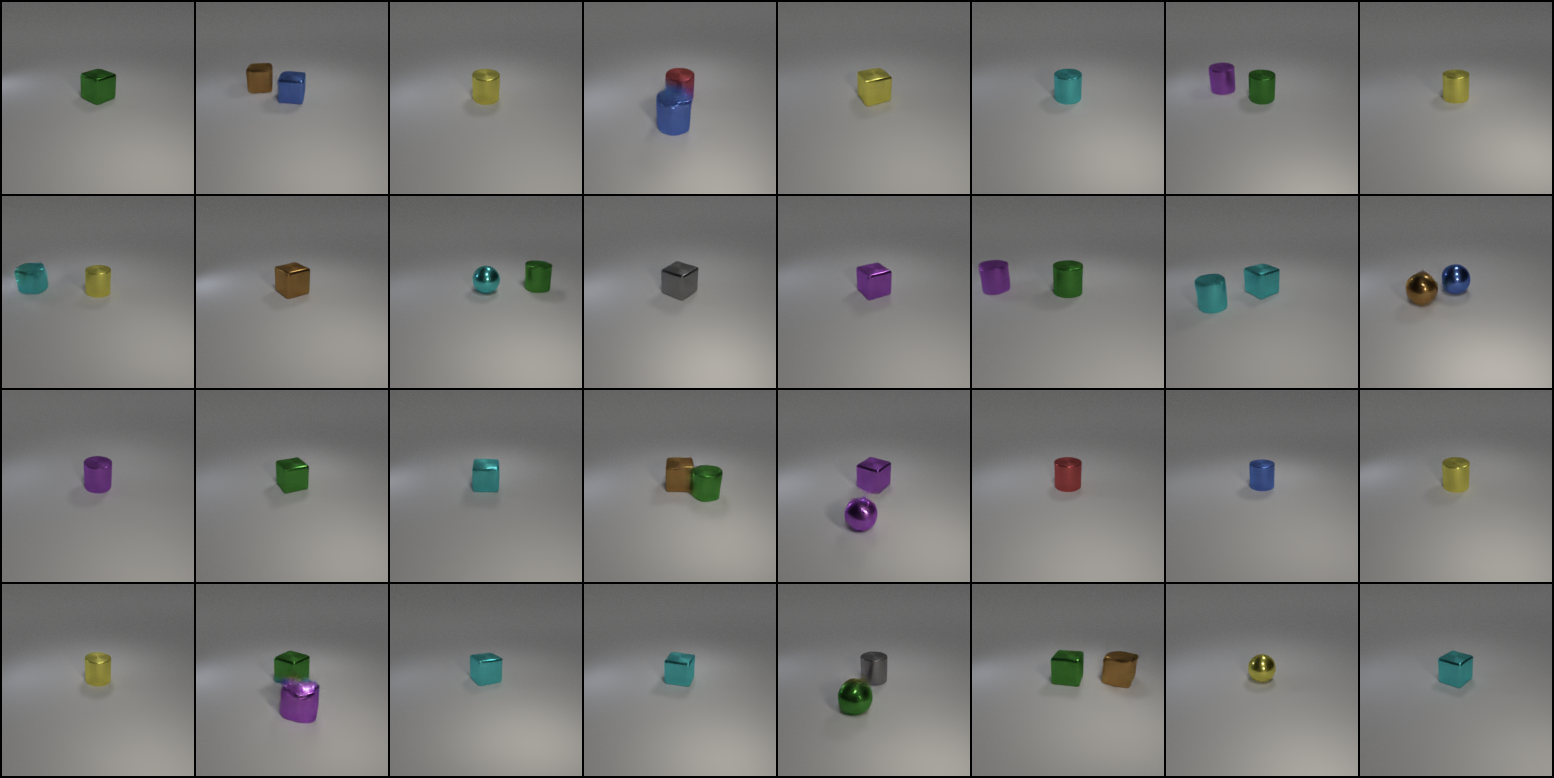}
    \caption{MSGNet CLEVR generations $192\times 192$.}
    \label{layoutgens}
\end{figure*}

The small size of the images sometimes made it difficult to distinguish some generated objects from each other, e.g., a red cube and a red cylinder, therefore we synthesized \footnote{We refer to  \href{https://github.com/facebookresearch/clevr-dataset-gen}{https://github.com/facebookresearch/clevr-dataset-gen} for the CLEVR data creation process.} $20,000$ new images of size $192\times 192$. In contrast to the former CLEVR dataset, we allowed only 1-2 objects within an image. Every image had one object fixed in the middle, while an optional second object circulated the middle object. This creates a reflective semantic dependency: reflections may only be visible when the second object is near the object in the middle.  \autoref{layoutgens} further demonstrates that MSGNet can capture the semantic and spatial dependencies between the objects. Since we are providing a layout as conditioning, we can not only control the type and spatial location but also the size of the object. 

\subsection{Malaria}\label{malaria-appendix}
\begin{figure*}
    \centering
    \includegraphics[scale=0.5]{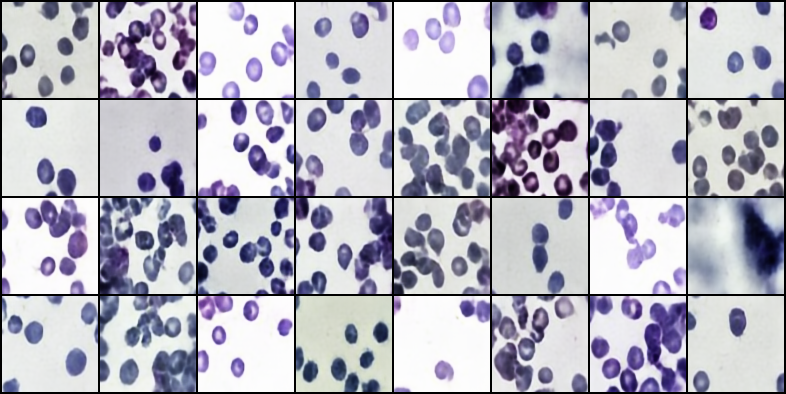}
    \caption{MSGNet Malaria cell generation.}
    \label{malaria_generations}
\end{figure*}

The data consisted of two uninfected classes and four other classes indicating different stages of the malaria-infection process. In this study we merged all the infected classes into one malaria-infected class, the other classes along with the background were treated as one class. A class imbalance existed between healthy and infected cells: the healthy class made up over 95\% of all cells. Each cell was accompanied by a label and bounding box coordinates, which we used to make the corresponding layout for the extracted patch. From the down-scaled 1208 training images, we extracted 10,000 patches of size $96\times 96$.
We further augmented this dataset with 6500 generated patches. In \autoref{malaria_generations}, we can see generated malaria patches.

\subsection{BraTS}\label{brats-appendix}

\begin{figure*}[h!]
\begin{subfigure}{\textwidth}
\centering
\caption{Real samples}
\includegraphics[scale=0.26]{images/brats_tests/real_examples/samples/1orig.jpg}
\includegraphics[scale=0.26]{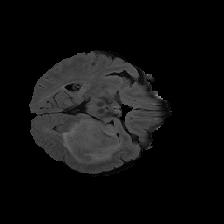}
\includegraphics[scale=0.26]{images/brats_tests/real_examples/samples/4orig.jpg}
\includegraphics[scale=0.26]{images/brats_tests/real_examples/overlays/1mask.jpg}
\includegraphics[scale=0.26]{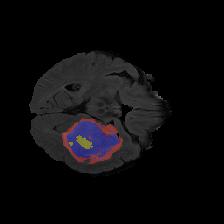}
\includegraphics[scale=0.26]{images/brats_tests/real_examples/overlays/4mask.jpg}
\label{fig_brain_real}
\end{subfigure}

\begin{subfigure}{\textwidth}
\centering
\caption{MSGNet generations}
\includegraphics[scale=0.26]{images/brats_tests/gen_examples/samples/unet_true26nomask.jpg}
\includegraphics[scale=0.26]{images/supplementary/brats/unet_true3nomask.jpg}
\includegraphics[scale=0.26]{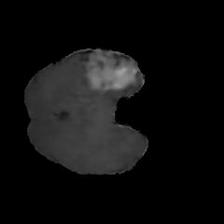}
\includegraphics[scale=0.26]{images/brats_tests/gen_examples/overlays/unet_true26.jpg}
\includegraphics[scale=0.26]{images/supplementary/brats/unet_true3.jpg}
\includegraphics[scale=0.26]{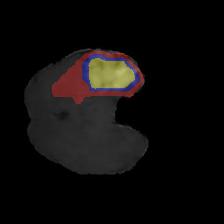}
\end{subfigure}

\begin{subfigure}{\textwidth}
\centering
 \includegraphics[scale=0.26]{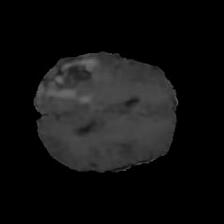}
 \includegraphics[scale=0.26]{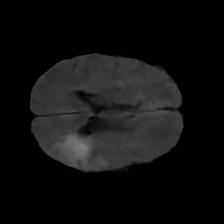}
  \includegraphics[scale=0.26]{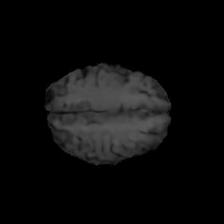}
   \includegraphics[scale=0.26]{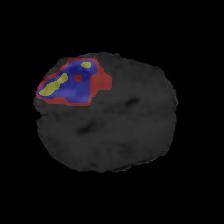}
 \includegraphics[scale=0.26]{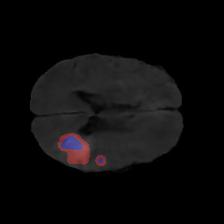}
  \includegraphics[scale=0.26]{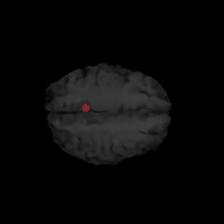}
  \end{subfigure}

\begin{subfigure}{\textwidth}
\centering
 \includegraphics[scale=0.243]{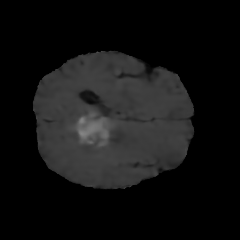}
 \includegraphics[scale=0.243]{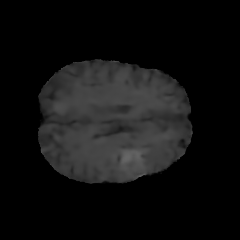}
  \includegraphics[scale=0.243]{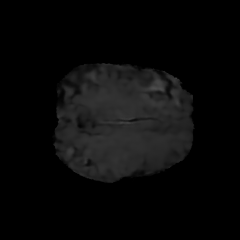}
   \includegraphics[scale=0.243]{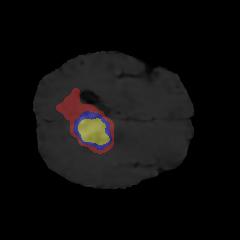}
 \includegraphics[scale=0.243]{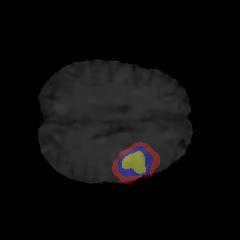}
  \includegraphics[scale=0.243]{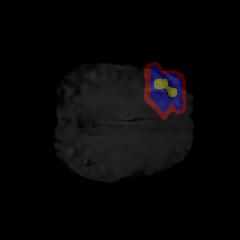}
 \end{subfigure}

\caption{Image slices of brain (top) and the respective overlaid tumor masks (bottom), Red: peritumoral edema, Yellow: necrotic and non-enhancing tumor core, Blue: enhancing tumor. The first row shows real samples with their corresponding layouts, while the lower two rows show six generations with the generated layouts.}
\label{brats_mask_generated_image}
\end{figure*}

A subset of the BrATS dataset was used. We used the $250\times 240\times 155$ images accompanied by segmentation layouts from the FLAIR and T1Gd modes. The segmentation layouts showed the sub-regions for three different classes:  GD-enhancing tumor, the peritumoral edema, and the necrotic and non-enhancing tumor core. The brain and the background were treated as an additional class. We refer to \cite{Menze2015tmi} for further details. For the two modes, the background has been modelled using two different classes. We only considered half of the 155 slices, i.e., we took only the even-numbered slices and out of those we discarded the slices that did not include one of the previously mentioned tumor-classes in order to create a more balanced dataset. This resulted in 20,000 images that were further augmented with 6,000 MSGNet generated images. Example generations of the masks as well as the images are shown in \autoref{brats_mask_generated_image}.

\begin{figure*}[h!]
\begin{subfigure}{\textwidth}
    \centering
 \includegraphics[scale=0.7]{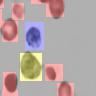}
 \includegraphics[scale=0.7]{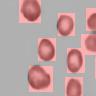}
 \includegraphics[scale=0.7]{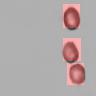}
  \includegraphics[scale=0.7]{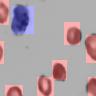}
  \includegraphics[scale=0.7]{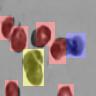}
     \caption{Masks}
 \end{subfigure}
 \begin{subfigure}{\textwidth}
    \centering
 \includegraphics[scale=0.7]{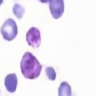}
 \includegraphics[scale=0.7]{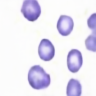}
 \includegraphics[scale=0.7]{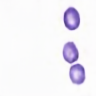}
  \includegraphics[scale=0.7]{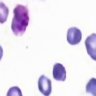}
  \includegraphics[scale=0.7]{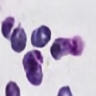}
      \caption{Generated Images}
 \end{subfigure}
    \caption{MSGNet generations. The generated masks (a) overlaid on the generated cells in (b).}
    \label{mask_generated_image}
\end{figure*}

\subsection{CityScape}\label{cityscape-appendix}

\begin{figure*}[h!] 
    \centering
 \includegraphics[scale=0.7]{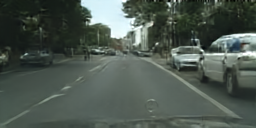}
 \includegraphics[scale=0.7]{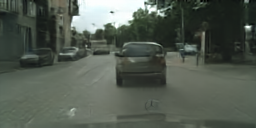}
 \includegraphics[scale=0.7]{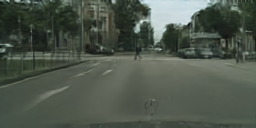}
  \includegraphics[scale=0.7]{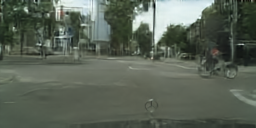}
  \includegraphics[scale=0.7]{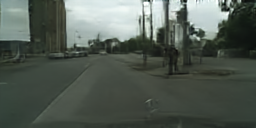}
 \includegraphics[scale=0.7]{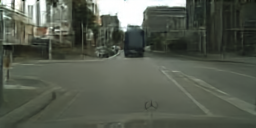}
 \includegraphics[scale=0.7]{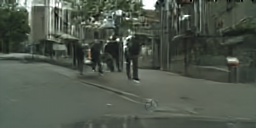}
  \includegraphics[scale=0.7]{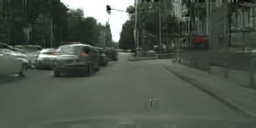}
  \includegraphics[scale=0.7]{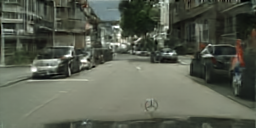}
  \includegraphics[scale=0.7]{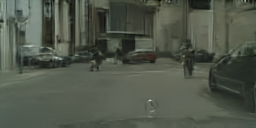}
    \caption{MSGNet generations on the cityscape 256$\times$128 dataset.}
    \label{fig:cityscape}
\end{figure*}

We performed a preliminary study on the cityscape dataset \cite{cityscape}. We used $256\times 128$ sized images with a network latent size of $64\times 32$. We only trained the VQ-VAE and the PixelSNAIL to generate the image encodings, the LayoutPixelSNAIL was not used in this case. In \autoref{fig:cityscape} we show ten generated images.

\newpage
\section{Architecture Details}\label{arch-appendix}
\begin{figure*}[h!] 
    \centering
    \includegraphics[width=\textwidth]{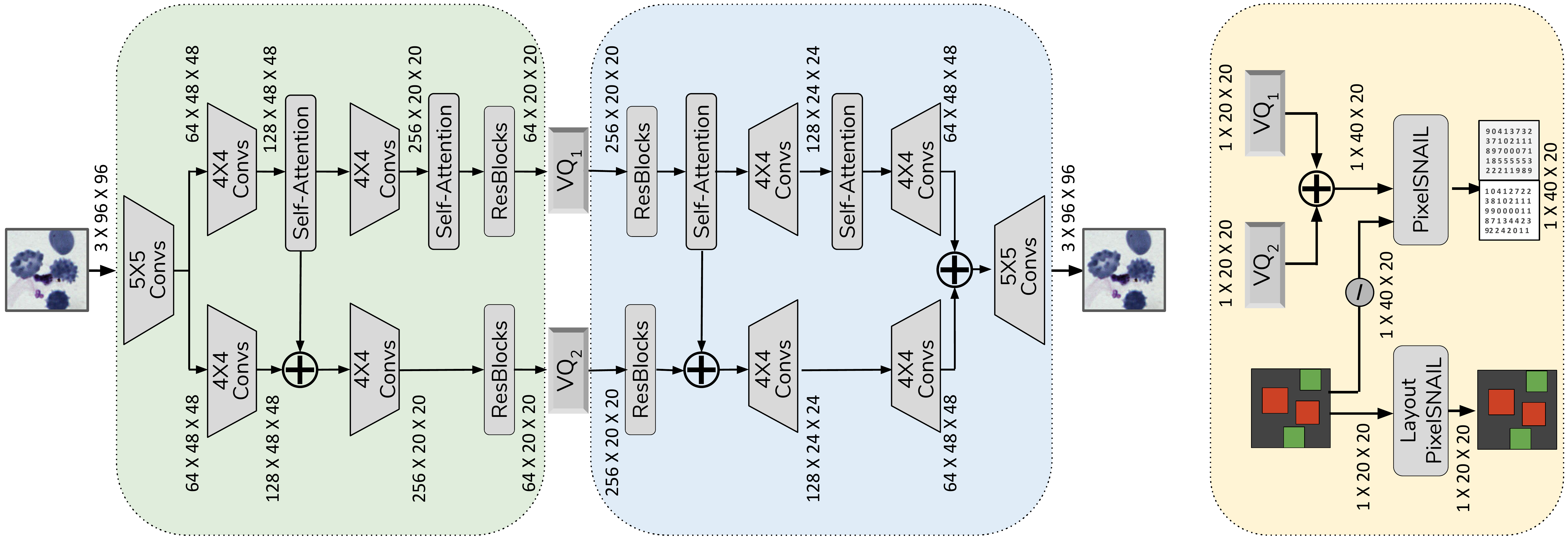}
    \caption{The VQ-VAE and PixelSNAIL training process with the output shapes for the intermediary layers. In our experiments the size of the prototype-vectors in the codebook was 64. These vectors have a corresponding index, the distribution of which are learned during the PixelSNAIL latent training phase.}
    \label{framework_dimension}
\end{figure*}
In this section, we provide further details on the architectures used for the CLEVR, Malaria and BraTS datasets. The schematic of  \autoref{framework_dimension} largely shows the baseline double-path VQ-VAE architecture that we used along with the dimensions at each layer. 

Tables \ref{vqvaehp}, \ref{PixelSNAILHP} and \ref{hyperparamslayout} provide hyperparameter details for all the datasets with the respective models. Moreover, in all the PixelSNAIL training procedures we used a cyclical learning rate \cite{cyclicallr2017leslie}. In all our experiments, both the encoder and decoder of the VQ-VAE used strided (transposed-) convolutions to increase or decrease the width and height of the features. We used the self-attention modules as seen in \cite{zhang2019icml}. 

We note that because of the varying image dimensions of the datasets, the placement of the self-attention modules might differ. That is, to accommodate for computational overhead brought by the inclusion of the self-attention modules, we decided to add this module only to feature spaces with width and height smaller than 64. For example, The self-attention modules were added after the first strided convolution in the CLEVR-64 experiments, since both width and height were halved in size after one convolution. In addition, concatenating the output of the self-attention module to the normal-path can be done at different stages in the encoding and decoding forward pass. We mainly found that adding a second path that provided complementary information to the normal-path sufficed to lower the reconstruction error.

\begin{table*}[ht]
\centering
\begin{tabular}{l|llll}
                & CLEVR-64         & CLEVR-192        & Malaria          & BraTS            \\ \hline
Image size      & $64\times 64$    & $192\times 192$  & $96\times 96$    & $240\times 240$  \\
Batch size      & 64               & 32               & 32               & 16               \\
Hidden dim      & 128              & 128              & 256              & 256              \\
Residual dim    & 64               & 64               & 64               & 64               \\
Residual blocks & 2                & 2                & 2                & 2                \\
Codebook size   & 64               & 64               & 64               & 64               \\
Codebook num    & 256              & 256              & 128              & 128              \\
Latent size     & 16               & 24               & 20               & 30               \\
Commitment      & 0.25             & 0.25             & 0.25             & 0.25             \\
Learning rate   & $1\cdot 10^{-3}$ & $8\cdot 10^{-4}$ & $8\cdot 10^{-4}$ & $8\cdot 10^{-4}$ \\
Scheduler       & Linear           & Linear           & Linear           & Linear           \\
Optimizer       & Adam             & Adam             & Adam             & Adam             \\
Iterations      & 25000            & 50000            & 80000            & 50000           
\end{tabular}
\caption{Hyper parameters for both paths in the VQ-VAE encoder and decoder. }
\label{vqvaehp}
\end{table*}

\begin{table*}[ht]
\centering
\begin{tabular}{l|llll}
                            & CLEVR-64         & CLEVR-192        & Malaria          & BraTS            \\ \hline
Latent size                 & $32\times 16$    & $48\times 24$    & $40\times 20$    & $60\times 30$    \\
Batch size                  & 32               & 16               & 16               & 16               \\
Hidden dim                  & 256              & 128              & 128              & 256              \\
Residual dim                & 256              & 128              & 128              & 256              \\
Residual blocks             & 3                & 3                & 3                & 3                \\
Output residual blocks      & 0                & 0                & 0                & 0                \\
Conditional residual blocks & 2                & 2                & 2                & 2                \\
Conditional residual dim    & 64               & 64               & 64               & 128              \\
Condition Layout  dim       & 64               & 128              & 32               & 64               \\
Attention dim               & 128              & 64               & 64               & 128              \\
Attention heads             & 8                & 8                & 8                & 8                \\
dropout                     & 0.1              & 0.15             & 0.15             & 0.15             \\
Learning rate               & $3\cdot 10^{-4}$ & $3\cdot 10^{-4}$ & $3\cdot 10^{-4}$ & $3\cdot 10^{-4}$ \\
Scheduler                   & Cyclical         & Cyclical         & Cyclical         & Cyclical         \\
Optimizer                   & Adam             & Adam             & Adam             & Adam             \\
Epochs                      & 60               & 75               & 50               & 75              
\end{tabular}
\caption{The hyperparameters used for each dataset in the PixelSNAIL model.}
\label{PixelSNAILHP}

\end{table*}
\begin{table*}[h!]
\centering
\begin{tabular}{l|llll}
                            & CLEVR-64         & CLEVR-192        & Malaria          & BraTS            \\ \hline
Layout size                 & $16\times 16$    & $24\times 24$    & $20\times 20$    & $30\times 30$    \\
Batch size                  & 64               & 32               & 32               & 32               \\
Hidden dim                  & 64               & 128              & 128              & 128              \\
Residual dim                & 64               & 128              & 128              & 128              \\
Residual blocks             & 2                & 2                & 3                & 3                \\
Output residual blocks      & 0                & 0                & 0                & 0                \\
Conditional residual blocks & -                & -                & -                & -                \\
Conditional residual dim    & -                & -                & -                & -                \\
Layout embedding  dim       & 64               & 32               & 64               & 64               \\
Attention dim               & 32               & 64               & 64               & 64               \\
Attention heads             & 8                & 8                & 8                & 8                \\
dropout                     & 0.15             & 0.15             & 0.15             & 0.15             \\
Learning rate               & $3\cdot 10^{-4}$ & $1\cdot 10^{-4}$ & $1\cdot 10^{-4}$ & $1\cdot 10^{-4}$ \\
Scheduler                   & Linear           & Linear           & Linear           & Linear           \\
Optimizer                   & Adam             & Adam             & Adam             & Adam             \\
Epochs                      & 20               & 15               & 15               & 20              
\end{tabular}
\caption{The hyperparameters used for the Layout PixelSNAIL}
\label{hyperparamslayout}
\end{table*}

\end{document}